\documentclass[letterpaper]{article} 
\usepackage{aaai24}  
\usepackage{times}  
\usepackage{helvet}  
\usepackage{courier}  
\usepackage[hyphens]{url}  
\usepackage{graphicx} 
\usepackage{natbib}  
\usepackage{caption} 
\usepackage{algorithm}
\usepackage{algorithmic}

\UseRawInputEncoding

\usepackage{multicol}
\usepackage{multirow}
\usepackage{booktabs}
\usepackage{amsmath}
\usepackage{breqn}
\usepackage{pgfplots}
\pgfplotsset{width=7cm,compat=1.18}
\usepackage{tikz}
\usepackage{xr}
\usepackage{mathtools}
\usetikzlibrary{intersections}
\usepackage{relsize}
\usepgfplotslibrary{fillbetween}
\usepackage{pgfkeys}
\newcommand{\Hquad}{\hspace{0.5em}}

\pgfplotsset{
        no markers,
        xmin=0,
        enlarge x limits=false,
        scaled y ticks=false,
        ymin=-1, ymax=1
}
\tikzset{line join=bevel}
\usepackage{color,xcolor,soul}

\usepackage{pgfplots, pgfplotstable}
\usepackage{color, colortbl}
\usetikzlibrary{shapes.geometric}
\usetikzlibrary{spy}
\usepackage{graphicx}
\usepackage[T1]{fontenc}
\definecolor{color1}{rgb}{0.392,0.392,1.0}
\definecolor{green}{rgb}{0.0,0.5,0.0}
\definecolor{red}{rgb}{1.0,0.01,0.24}
    \def\addlegendimage{\csname pgfplots@addlegendimage\endcsname}
\pgfplotsset{compat=1.11,
    /pgfplots/ybar legend/.style={
    /pgfplots/legend image code/.code={%
       \draw[##1,/tikz/.cd,yshift=-0.5em]
        (1cm,1cm) rectangle (5pt,0.8em);},
   },
}
\usepackage{pgfplotstable}

\usepackage[capitalize]{cleveref}
\crefname{section}{Section}{Sections}
\crefname{table}{Table}{Tables}
\crefname{table}{Table}{Tables}
\crefname{figure}{Figure}{Figures}
\Crefname{figure}{Figure}{Figures}
\pgfmathdeclarefunction{invgauss}{2}{%
  \pgfmathparse{sqrt(-2*ln(#1))*cos(deg(2*pi*#2))}%
}
\makeatletter
\pgfplotsset{
    table/.cd,
    brownian motion/.style={
        create on use/brown/.style={
            create col/expr accum={
                (\coordindex>0)*(
                    max(
                        min(
                            invgauss(rnd,rnd)*0.1+\pgfmathaccuma,
                            \pgfplots@brownian@max
                        ),
                        \pgfplots@brownian@min
                    )
                ) + (\coordindex<1)*\pgfplots@brownian@start
            }{\pgfplots@brownian@start}
        },
        y=brown, x expr={\coordindex},
        brownian motion/.cd,
        #1,
        /.cd
    },
    brownian motion/.cd,
            min/.store in=\pgfplots@brownian@min,
        min=-inf,
            max/.store in=\pgfplots@brownian@max,
            max=inf,
            start/.store in=\pgfplots@brownian@start,
        start=0
}

\usepackage{newfloat}
\usepackage{listings}
\DeclareCaptionStyle{ruled}{labelfont=normalfont,labelsep=colon,strut=off} 
\floatstyle{ruled}
\newfloat{listing}{tb}{lst}{}
\floatname{listing}{Listing}
\title{Contrastive Learning for Lane Detection via Cross-Cimilarity\footnote{The paper is accepted to publish at Pattern Recognition Letters}}
\author {
    Ali Zoljodi\textsuperscript{\rm 1},
    Sadegh Abadijou\textsuperscript{\rm 2},
    Mina Alibeigi\textsuperscript{\rm 3},
    Masoud Daneshtalab\textsuperscript{\rm 1,4}
}
\affiliations {
    \textsuperscript{\rm 1}M\"alardalen University\\
    \textsuperscript{\rm 2}University of Leicester\\
    \textsuperscript{\rm 3}Zenseact AB\\
    \textsuperscript{\rm 4}Tallinn University of Technology, Estonia\\
    ali.zoljodi@mdu.se, ma1023@leicester.ac.uk, mina.alibeigi@zenseact.com,
    masoud.daneshtalab@mdu.se
}

\usepackage{bibentry}

\begin{document}
\maketitle

\begin{abstract}

Detecting lane markings in road scenes poses a significant challenge due to their intricate nature, which is susceptible to unfavorable conditions. While lane markings have strong shape priors, their visibility is easily compromised by varying lighting conditions, adverse weather, occlusions by other vehicles or pedestrians, road plane changes, and fading of colors over time. The detection process is further complicated by the presence of several lane shapes and natural variations, necessitating large amounts of high-quality and diverse data to train a robust lane detection model capable of handling various real-world scenarios.

In this paper, we present a novel self-supervised learning method termed Contrastive Learning for Lane Detection via Cross-Similarity (CLLD) to enhance the resilience and effectiveness of lane detection models in real-world scenarios, particularly when the visibility of lane markings are compromised.
CLLD introduces a novel contrastive learning (CL) method that assesses the similarity of local features within the global context of the input image. It uses the surrounding information to predict lane markings.
This is achieved by integrating local feature contrastive learning with our newly proposed operation, dubbed \textit{cross-similarity}.
 
The local feature CL concentrates on extracting features from small patches, a necessity for accurately localizing lane segments. Meanwhile, cross-similarity captures global features, enabling the detection of obscured lane segments based on their surroundings. We enhance cross-similarity by randomly masking portions of input images in the process of augmentation.
Extensive experiments on TuSimple and CuLane benchmark datasets demonstrate that CLLD consistently outperforms state-of-the-art contrastive learning methods, particularly in visibility-impairing conditions like shadows, while it also delivers comparable results under normal conditions. When compared to supervised learning, CLLD still excels in challenging scenarios such as shadows and crowded scenes, which are common in real-world driving.

\end{abstract}
\section{Introduction}\label{section:intro}
Lane detection is a crucial task in computer vision, particularly for autonomous vehicles and advanced driver assistance systems. This process becomes even more challenging in diverse real-world scenarios, primarily because lane markings are inherently long, thin structures characterized by strong shape priors but limited appearance clues~\cite{Pan2018SpatialCNN}.
The visibility of these markings is frequently compromised by adverse factors such as poor lighting conditions, occlusions, and the fading of their color, all of which contribute to making lane detection a highly demanding task~\cite{Liu2021WACV}.
{Feature extraction is a crucial component of computer vision algorithms~\cite{fractalfract7080598,biology11121732}, especially for lane detection. However, it's not only important to extract features but also to capture the long-range dependencies between these features. This is essential for predicting lanes in segments where visibility is low.}
Many novel lane detection approaches adopt pixel-level image segmentation techniques~\cite{Zheng2021RESA,Pan2018SpatialCNN} to enhance the precision of lane detection.
In these methods, each pixel is labeled either as part of the lane or as a background.

\begin{figure}[ht]
    \centering
    \includegraphics[width=0.8\columnwidth]{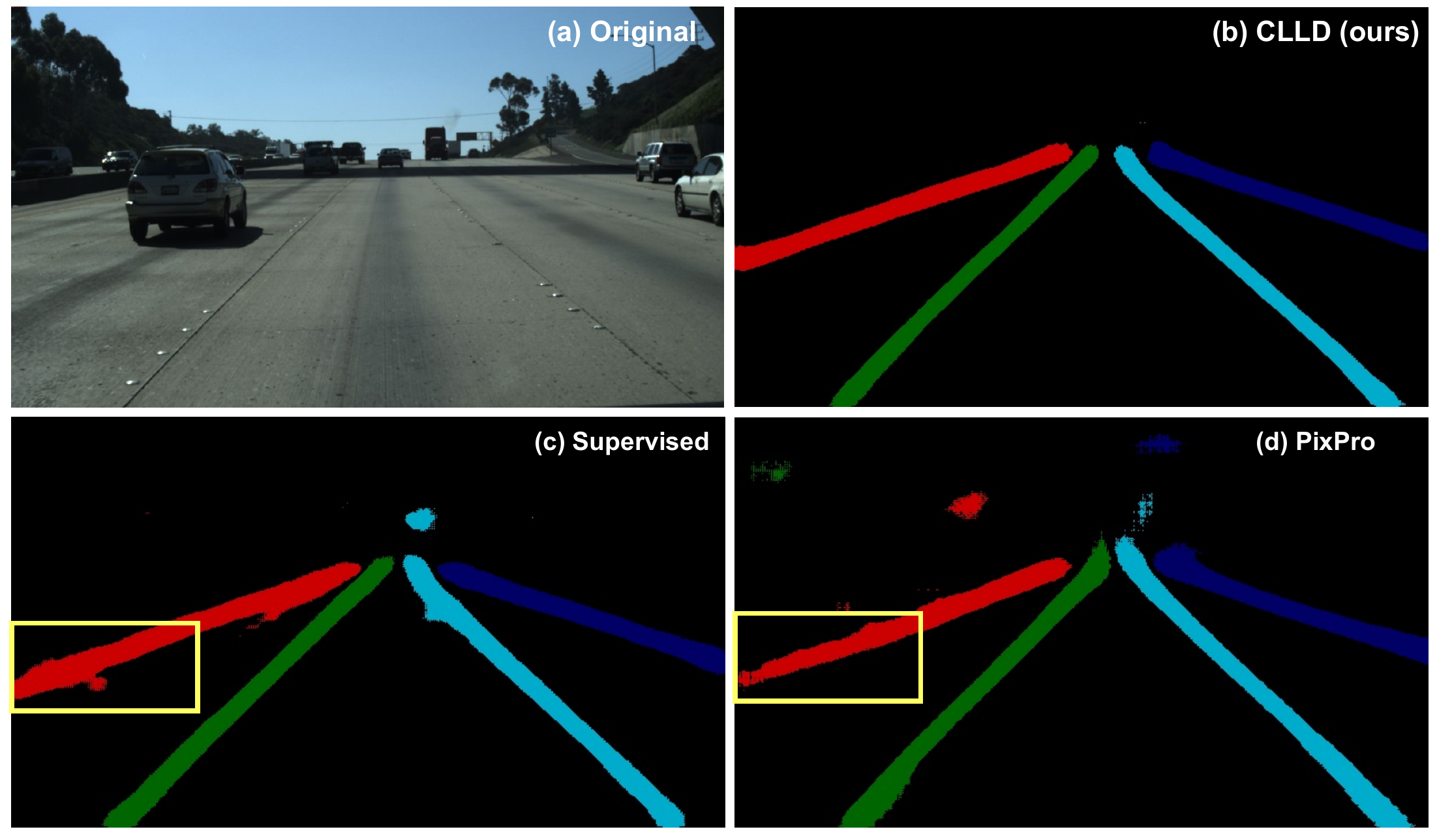}
    \caption{Comparison of the state-of-the-art segmentation-based lane detection RESA\cite{Zheng2021RESA} with three different pretraining strategies. (a) Input image (b) RESA output with  CLLD (ours) and (c) RESA output with supervised and (d) RESA output with PixPro\cite{Xie2021PropagateYourself} pretraining. Yellow boxes represent accuracy drops in the detection of lanes that are occluded by cars.}
    \label{fig:fig1}
\end{figure}  

To develop a robust lane detection method capable of handling natural variations, a significant amount of training data is required.
Large-scale labeling of lane markings in road scenes is costly and requires a lot of human labor. However, an abundance of unlabeled data is available, which can simply be used to boost the performance of the model.
Furthermore, the appearance of lanes varies across the globe.
Unsupervised and self-supervised \textit{contrastive learning} (CL) methods have been proposed for training \textit{deep neural networks} (DNNs) with minimal labeled data and a vast amount of unlabeled data~\cite{chen2020simclr,NEURIPS2022_39cee562}.

CL methods can be divided into two categories based on the type of representations or features they focus on in an image: local feature methods and global feature methods. Global feature CL methods are not considered the most effective strategy for lane detection, considering they compare the entirety of an input image with other images. However, for lane detection, it's crucial to localize specific lane segments within the same input image by contrasting different parts of it. On the other hand, local feature CL methods are designed to learn features that classify smaller portions of the input image. These methods can be effectively employed in object localization tasks like lane detection since they focus on specific areas within the input image, enabling more precise detection and localization of objects like lane lines \cite{NEURIPS2022_39cee562}.

Existing local CLs are not adequate for lane detection.
They are designed to detect completely visible objects and do not have any mechanism to predict the existence of objects that are obscured due to natural variations or occluding by vehicles and pedestrians.
To effectively detect and pinpoint lane markings, even in low visibility areas, we propose \textit{contrastive learning for lane detection via cross-similarity} (CLLD), a self-supervised learning method for lane detection.
CLLD is a multi-task contrastive learning approach. It trains \textit{convolutional neural networks} (CNNs) to segment an input image and predict masked parts of an image using their surrounding parts.
To train the model, we provide both an input image and its augmented version to an encoder. We then measure the consistency between the original image's feature map and the feature map generated from the augmented version that warped to its original shape.


{The utilization of masking to self-supervised learning models is referenced in notable works~\cite{pmlr-v162-shi22d}, which propose the masking models for training robust image classification models. 
However, a noteworthy challenge arises when the masked area is sufficiently large, leading to the CNN's inability to perform effectively, as CNNs inherently possess strong inductive bias. In addressing this issue, we introduce a novel operation named cross-similarity, a lightweight operation designed to leverage the similarities between the feature maps surrounding the masked area and their corresponding feature map in the original image. This innovative approach mitigates the loss of important features, thereby enhancing the CNN's capability to effectively detect and reconstruct objects even in scenarios characterized by significant occlusion.}
We compute the cross-similarity of every patch of the feature map in the masked image with all parts of the feature map in the original image and contrast it with the cross-similarity of each patch in the original image feature maps, and the entire masked image features maps.
{Differing from \cite{Shen_2023,SONG2024107945}, which mask data to reduce size and focus attention, CLLD uses masking to augment data and train a model to predict missing parts, to improve resiliency against occlusion and data loss.

We assess CLLD on U-Net~\cite{Ronneberger2015UNet}, a popular encoder-decoder that is widely utilized for lane detection tasks.
In addition to U-Net, we evaluate the proposed method on RESA \cite{Zheng2021RESA} and CLRNet \cite{Zheng_2022_CVPR} as SOTA segmentation-based and anchor-based lane detection methods, respectively.
CLLD yields an average $1\%$ improvement in all evaluation metrics over state-of-the-art CL methods on two of the most well-known lane detection datasets, CuLane~\cite{Pan2018SpatialCNN} and TuSimple~\cite{Shirke2019LaneDatasets}.
To demonstrate its efficacy, we have tested CLLD on the shadow subset of CuLane, which is known to be a challenging set due to its varying light conditions. Our findings show an impressive over 4\% improvement in detecting lanes in shadow situations.
From the qualitative results shown in Figure~\ref{fig:fig1}, we can see that CLLD outperforms SOTA local CL and supervised learning. Specifically, CLLD is more effective at dealing with occluded parts of a lane.

The main contributions of this work are as follows: \textbf{(I)} We demonstrate that previous self-supervised learning approaches may not be the most effective approach for the lane detection task. We highlight some reasons that may contribute to this performance reduction. \textbf{(II)} We present the cross-similarity approach, a lightweight operation that computes the correlation between spatial parts of a picture that may contain lane markings and connects them together. \textbf{(III)} We propose a novel approach to self-supervised learning for lane detection. Our approach leverages cross-similarity to pretrain lane detection to better detect occluded or worn-out lane segments. \textbf{(IV)} We show that our method surpasses supervised learning in detecting lanes under challenging conditions, like shadow-covered markings, by comparing CLLD's performance with that of supervised learning. \textbf{(V)} We demonstrate how our method can outperform supervised learning for lane detection in challenging scenarios, such as lane markings concealed by shadows, by comparing CLLD performance with supervised learning.
\section{Related work}\label{section:related_work}
\subsection{lane detection}\label{section:related_work:lane}
Lane detection is a critical module for autonomous driving.
The safety of autonomous vehicles is greatly affected by the accuracy and latency of lane detection methods.
Lane detection methods are classified as conventional~\cite{1707380} or based on CNN~\cite{Pan2018SpatialCNN,Zheng2021RESA,Zheng_2022_CVPR}.
Conventional lane detection~\cite{1707380} relies on manual features to identify lanes, limiting their accuracy in different road scenarios. {To detect lane segments, \cite{KANG20033177} combines local line extraction with dynamic programming to enhance the performance of Hough Transforms, which often struggle in complex scenes.}
{\cite{https://doi.org/10.1049/cth2.12441}  utilizes advanced feature extraction methods, such as LGBPHS and MTP, and optimizes the classification process with a BI-GRU, enhanced by Self-Improved Honey Badger Optimization, to accurately identify lane lines under various conditions, a technique that parallels our focus on robust lane detection.}

CNN technology has enabled new solutions for lane detection, such as U-Net~\cite{Tran2019LaneMarkings}. However, challenges arise when detecting occluded lanes due to biases and spatial information capture limitations. Spatial CNN~\cite{Pan2018SpatialCNN} and Recurrent Feature-Shift Aggregator~\cite{Zheng2021RESA} address these challenges by using message-passing to propagate spatial information. Anchor-based~\cite{Zheng_2022_CVPR} lane detection options are also some lane detection methods that behave lanes as a chain of anchors.
{3D lane detection offers an alternative approach to solving lane detection challenges by using road curvature patterns and aligning lane markings to these patterns. \cite{s23115358} propose a novel 3D lane detection method that utilizes improved feature extraction techniques and optimized BI-GRU classifiers.}
This paper discusses how CLLD can enhance the accuracy and robustness of segmentation-based and anchor-based lane detection. 
\subsection{contrastive learning}\label{section:related_work:CL}
\subsubsection{global features}
%
Global feature methods can aid in image classification by comparing positive and negative samples. Studies such as \cite{Tian2020ContrastiveCoding}, SimCLR ~\cite{chen2020simclr}, and MoCo~\cite{He2020CVPR} use different techniques to train the network to produce similar representations for all views of a sample. However, these methods may not work well for identifying specific parts of an image, as noted by \cite{Xie2021PropagateYourself}. { To improve the quality of self-supervised learned feature representation, \cite{LI2023155} integrates Bregman divergence into contrastive learning to enhance the learning of distance features between the latent features in the embedding space.}

\subsubsection{local features}
Local features are proposed to overcome pixel-level and region-level classifications.
Different levels of local feature methods can be utilized, such as at the feature-level~\cite{Wang2021DenseCL,NEURIPS2022_39cee562}, pixel-level~\cite{Xie2021PropagateYourself,Zhao_2021_ICCV}, or region-level~\cite{Xiao2021RegionSimilarity}.
DenseCL~\cite{Wang2021DenseCL} is a method that discriminates at the feature-level, inspired by MoCo-V2~\cite{He2020CVPR}.
PixPro~\cite{Xie2021PropagateYourself} is a method that discriminates at the pixel-level, inspired by BYOL~\cite{NEURIPS2020_f3ada80d}.
The positive samples are identified by pixels with Euclidean distance smaller than a threshold.
VICRegL~\cite{NEURIPS2022_39cee562} is a trade-off of global and local features that aims to achieve a balance in representation learning.
Detecting obscured lanes can be difficult with local feature CLs as they only extract visible segments. However, our method, CLLD, is capable of not only extracting visible parts of lanes but also predicting the existence of obscured lanes based on their surrounding visible parts.

\section{Methodology}\label{section:method:problem}
CLLD is a self-supervised approach that enhances lane detection. It focuses on understanding relationships between different image patches, enabling the detection of occluded or low-visibility lane segments by analyzing their surroundings. This method effectively improves lane detection accuracy when lane markings are poorly visible.
Utilizing the concept of CLLD, lane detection encoders are pretrained to reconstruct less visible lane segments by solving \cref{Eq:1}.
\begin{equation}
    \label{Eq:1}
    \mathcal{W}^*=\arg\min_{\mathcal{W}} \sum_{p \in P}{\mathcal{L}}_{clld}(\mathcal{F}(\mathcal{I}_p),\mathcal{F}(\mathcal{M}(\mathcal{I}_p))
\end{equation}
The variable $\mathcal{W}^*$ denotes the optimized weights for the lane detection encoder, and $\mathcal{I}$ is the training input.
The optimization problem involves extracting local features by dividing the input $\mathcal{I}$ into $P$ patches. The contrastive loss $\mathcal{L}(.)$ is applied to each patch $p\in P$, which undergoes two passes through the CNN backbone $\mathcal{F}$. One pass is in the original shape $\mathcal{F}(\mathcal{I}_p)$, and the other pass is in the masked shape $\mathcal{F}(\mathcal{M}(\mathcal{I}_p))$.
The objective of the optimization problem is to minimize the summation of loss values for all patches, denoted as $\sum_{p \in P}(.)$.

We consider lanes as objects with strong shape priors, adhering to a consistent pattern, yet occasionally exhibiting invisibility in random sections.
Empirical studies have shown that masking specific areas of the input image and training the encoder to predict those parts can be a beneficial method for teaching the lane detection backbone to predict occluded or missing lane segments. Consequently, we adopt this technique by masking the input image to generate a second view, which is then employed by the contrastive learning method.
Below, we provide a comprehensive explanation of the masking that we employed.

In local feature CL, after extracting features, the method wraps back augmentations to position the features in their original location on the image. This technique enables a direct comparison between the extracted local features and their original counterparts. Such a comparison is essential for learning the variations among features from different segments of the same image, thereby enhancing the method's efficacy in feature analysis and interpretation.
When using masking, it is not possible to warp areas that have been masked, as no features are extracted from those sections.
Our proposal method involves a crucial step to overcome the aforementioned difficulty - enhancing the extracted feature maps with \texttt{cross-similarity} information. 
This module calculates the similarity between each patch of the feature map from the original image, denoted as $\mathcal{F}({\mathcal{I}}_P)$, and all patches from the feature map of the masked image, denoted as $\{\forall p \in P: \mathcal{F}({\mathcal{M}}({\mathcal{I}}_p)\}$. It also compares each patch of the feature map from the masked image, $\mathcal{F}({\mathcal{M}}({\mathcal{I}}_p)\}$, to all patches from the feature map of the original image, denoted as $\{\forall p \in P: \mathcal{F}({\mathcal{I}}_p\})$.

Through \texttt{cross-similarity}, each patch $\mathcal{F}(\mathcal{I}_P)$ has the ability to interact with all the patches in the cross-view, thereby maintaining the positional information of the patches.
\texttt{cross-similarity} allows for the use of the local feature CL on masked images. This is done by sharing information from each patch with all the other patches in the cross-view, particularly the corresponding patch.
We provide a comprehensive explanation of the \texttt{cross-similarity}.
\subsection{Contrastive Learning for Lane Detection}\label{section:Method:CLLD}
Our method (CLLD) employs momentum contrastive learning~\cite{Xie2021PropagateYourself} (\ref{fig:2}).
Given the input image $\mathcal{I}$, the masking function masks some patches with the size $\rho \times \rho$ producing a masked view of the input, denoted as $\mathcal{M}(\mathcal{I})$.

We input $\mathcal{I}$ and $\mathcal{M}(\mathcal{I})$ into two different encoders.
The first encoder updates its weights using gradient descent, while the second one updates its weights using the momentum of the first encoder.
The output of $\mathcal{M}(\mathcal{I})$ is a feature map $\mathcal{F}(\mathcal{M})(\mathcal{I})$ that may not contain valid information for the masked areas.
To enhance the masked areas with comparable information, we compute the \texttt{cross-similarity} of each patch from $\mathcal{F}(\mathcal{M}(\mathcal{I}))$ and the entire feature map of the input $\mathcal{F}(\mathcal{I})$ . 
We perform the inverse operation on each patch within $\mathcal{F}(\mathcal{I})$ and the entirety of $\mathcal{F}(\mathcal{M}(\mathcal{I}))$, as discussed above.
\begin{figure}[ht]
    \centering
    \includegraphics[width=0.7\columnwidth]{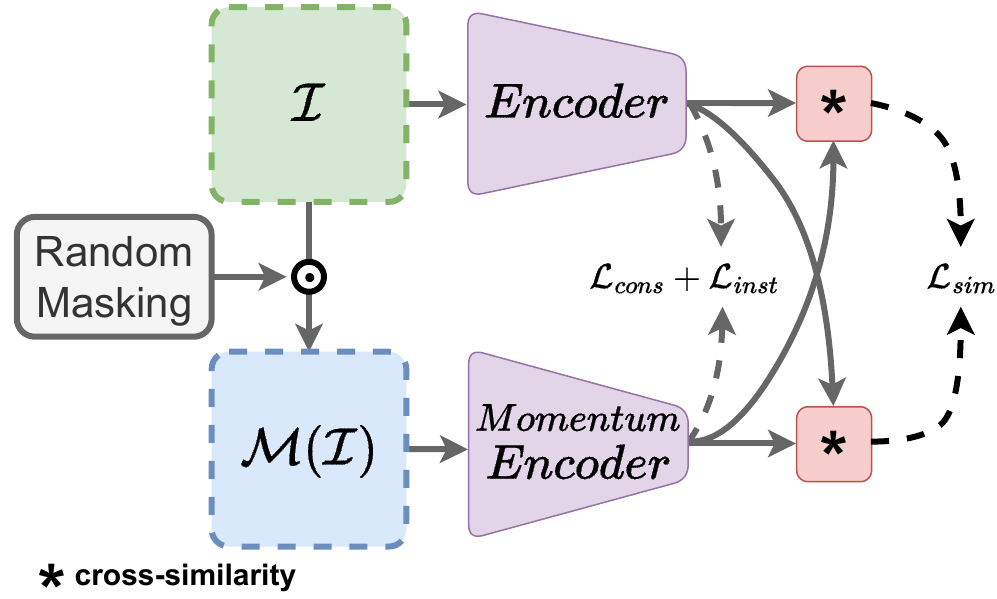}
    \caption{The CLLD framework.  }
    \label{fig:2}
\end{figure}

\subsection{Masking}\label{section:Method:Masking}
In this study, similar to masked image modeling approach ~\cite{pmlr-v162-shi22d}, we mask random portions of inputs and train the model to predict masked parts.
This approach trains lane detection backbones to predict hidden objects based on their surroundings an essential application for lane detection algorithms when dealing with occluded or vanished lane markings.
The random locations are square patches (with size $\rho\times\rho)$ of input images.
For a given input $\mathcal{I} \in [0,1]^{H\times W\times C}$ and portions to mask $\mathcal{M}=\{[0,1]^{i\times j}, where\Hquad i \subset \{0,H/P\}\Hquad and\Hquad  j \subset \{0,W/P\}\}$, the masked image is generated using \ref{Eq:2}.
\begin{equation}\label{Eq:2}
\begin{split}\scriptsize
  \scriptsize  \mathcal{M}(\mathcal{I})=\left\{
	\begin{array}{ll}
		\mathcal{N}(0,1)  & \mbox{if } \mathcal{M} \\
		\mathcal{I} & otherwise
	\end{array}
\right.
\end{split}
\end{equation}
Where $\mathcal{M}(\mathcal{I})$ is the masking image. 
To replace the pixel in a patch, we select a value from the normal distribution $\mathcal{N}(0,1)$. 

\begin{figure}[ht]

    \centering
    \includegraphics[width=\columnwidth ]{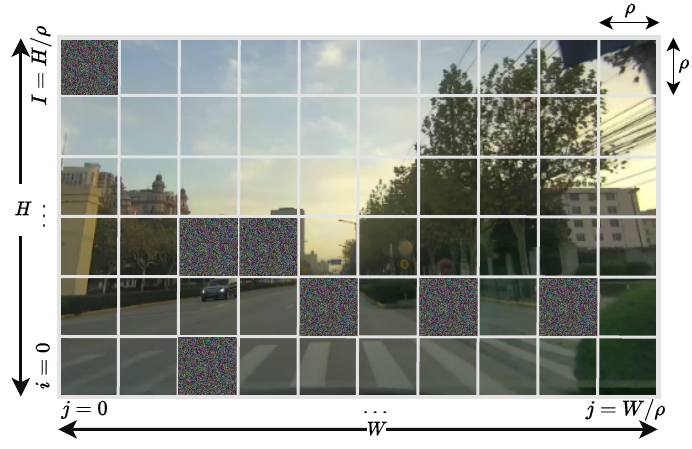}
    \caption{Masking input image; The given input with size $H \times W$ is divided into $\rho \times \rho$ patches. Each pixel of the masked patch got a random value from a zero-mean normal distribution $\mathcal{N}(0,1)$.} 
    \label{fig:3}
\end{figure}
\subsection{cross-similarity}\label{section:Method:Cross_sim}
%
To uncover the local features of masked patches, we utilize the similarity between their surrounding features and their corresponding features in the original image by applying \texttt{cross-similarity} operation~(\ref{fig:4}).

\begin{figure}[ht]
    \centering
    \includegraphics[width=\columnwidth]{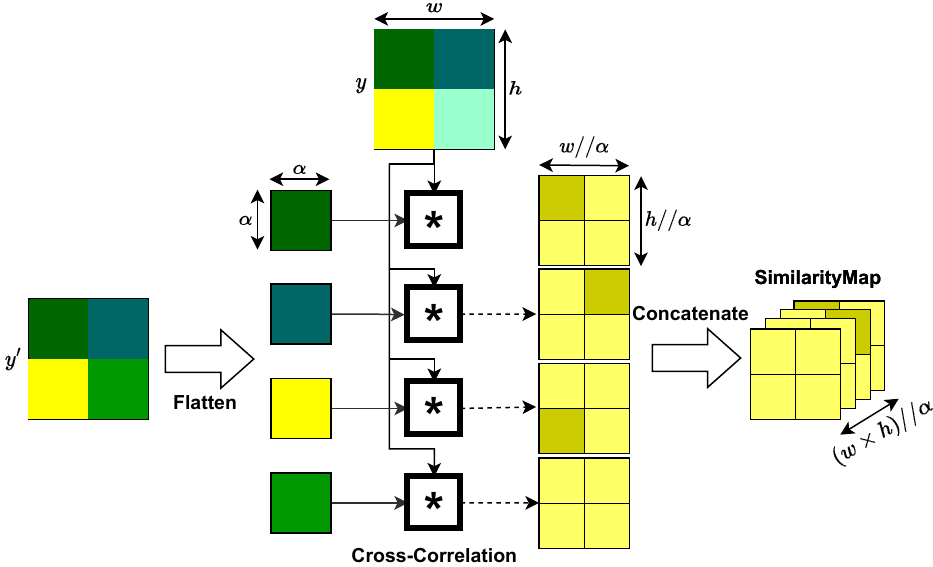}
    \caption{The cross-similarity operation}
    \label{fig:4}
%
\end{figure}

Assume that the output obtained from feeding the original input image $I$ is $y$ and the output obtained from feeding masked image $\mathcal{M}(I)$ is $y'$.
In order to extract local features, the contrastive loss needs to be applied to the small patches of $y$ and $y'$. These patches are denoted by $\{y_p\Hquad and\Hquad  {y'}_p, \forall p \in P\}$, where $P$ represents all patches within a feature map.
The \texttt{cross-similarity} between the full feature map of the first view and a patch of the feature map of the second view ${y'}_p$ is calculated through the use of \ref{Eq:3}.
\begin{align}
\label{Eq:3}
\begin{split}
CS(y,{y'}_p)[i,j]= \sum_{u=-\frac{\alpha}{2}}^{\frac{\alpha}{2}} \sum_{v=-\frac{\alpha}{2}}^{\frac{\alpha}{2}} {y'}_p[u,v]y[i+u,j+v]  \\  \forall i \in [0,h//\alpha] \Hquad and \Hquad \forall j \in [0,w//\alpha]
\end{split} 
\end{align}

The variable $\alpha$ represents the measurement of the sides of each patch.
Additionally, $h$ and  $w$ denote the height and width of feature maps, respectively.
To generate the complete \texttt{cross-similarity} between $y$ and $y'$, we compute the \texttt{cross-similarity} between $y$ and every patch of $y'$ ($\forall p \in y'$) and then concatenate them all together (\ref{Eq:4}).
\begin{align}
\label{Eq:4}
\begin{split}
CS(y,( \forall p \in y'))=&[CS(p_0,y),CS(p_1,y),...,CS(p_z,y)]  \\ &where \Hquad z=(w\times h)//\alpha
\end{split} 
\end{align}
In order to calculate the \texttt{cross-similarity} of $y$ with $y'$, denoted as $CS(y',( \forall p \in y))$, we perform the same operation as before but with $y$ and $y'$ swapped.

As shown in \ref{fig:4}, the \texttt{cross-similarity} of each patch ${y'}_p$ with the feature map $y$  generates a tensor. These tensors are then concatenated in their respective order.
By using this mechanism, one can not only discover the similarity between each patch from one feature map and all patches from the other but also retain their location information.
Therefore, the contrastive loss can reflect the patterns between the location and the similarity value to the neural network. 
Furthermore, the contrastive loss between the feature maps of the original view and the masked portions in the second view is higher than that of the other sections. Consequently, as the loss value for the masked sections increases, there is a corresponding increase in attention toward predicting these areas.
\subsection{Loss function}\label{section:Method:Loss}
Three key tasks are considered to design loss function.
\subsubsection{Consistency loss} 
The first objective of CLLD is to extract visible segments from the input.
To train the CNN backbone to extract accurate features, we utilize a $\mathcal{L}_{const}$ (\ref{Eq:5}), which contrasts the consistency of the feature maps produced by CNNs for two views (original and masked). 
\begin{equation}
\label{Eq:5}
  \mathcal{L}_{cons}=  -\frac{1}{h\times w} \times \sum_{i=1}^{h} \sum_{j=1}^{w} \frac{y_{ij}.{y'}_{ij}}{\parallel y_{ij}\parallel_2 \times \parallel {y'}_{ij}\parallel_2}  
\end{equation}
To evaluate their consistency, we compute the cosine similarity between each pixel on feature map $y$ and its corresponding pixel on feature map $y'$.
A positive cosine similarity value indicates that the features are consistent, implying a similar orientation in the feature space. Conversely, a negative value signifies inconsistency, suggesting that the features are oriented in opposite directions in the feature space.
Therefore, to ensure a positive loss value for inconsistencies, we multiply the cosine similarity value by a negative sign. This approach ensures that a higher loss corresponds to greater inconsistency.
Finally, we compute the average of the cosine similarities across all pixels in the feature maps. This average represents the overall consistency between the feature maps, providing a single metric that encapsulates the similarity of the entire feature space.
\subsubsection{Similarity loss}
Another key objective of CLLD approach is to train CNNs to accurately predict features for masked areas.
To achieve this objective, we introduce a similarity loss mechanism $\mathcal{L}_{sim}$. This mechanism is designed to quantify the difference between the predicted features of the masked areas and their actual features, thereby guiding the CNN to make more accurate predictions.

The \textit{similarity loss} (\ref{Eq:6}) computes the consistency between two sets of cross-similarities $CS(y,( \forall p \in y'))$ and $CS(y',( \forall p \in y))$.
\begin{equation}\label{Eq:6}
    \mathcal{L}_{sim}=\\{\tiny -\frac{CS(y,( \forall p \in y'))\quad . \quad CS(y',( \forall p \in y))}{\parallel CS(y,( \forall p \in y')) \parallel_2\times\parallel CS(y',( \forall p \in y))\parallel_2}}
\end{equation}
$\mathcal{L}_{sim}$ is the cosine similarity between two cross similarities. 
Differences in results for $CS(y,( \forall p \in y'))$ and $CS(y',( \forall p \in y))$ may arise due to the presence of masked areas on the feature map $y'$. By leveraging cosine similarity, we can detect variations between different patches (\ref{fig:4}). This detection enables us to train our model with $\mathcal{L}_{Sim}$,  focusing on predicting the masked areas by detecting sources of inconsistency.
\subsubsection{Classification loss}
To learn the categorization representation, we employ an instance-level cosine similarity loss $\mathcal{L}_{inst}$ (~\ref{Eq:7}).
\begin{equation}\label{Eq:7}
    \mathcal{L}_{inst}=2-2\frac{\hat{y}\quad.\quad\hat{y'}}{\parallel \hat{y} \parallel_2 \times \parallel \hat{y'} \parallel_2}
\end{equation}
where the $\hat{y}$ and $\hat{y'}$ are normalized vectors of $y$ and $y'$, respectively.
The loss function we used in this study is the summation of all aforementioned loss functions (\ref{Eq:8}).
\begin{equation}\label{Eq:8}
    \mathcal{L}_{clld}=\mathcal{L}_{cons}+\mathcal{L}_{sim}+\mathcal{L}_{inst}
\end{equation}
\section{Experimental setup}\label{section:Experimental}
{For a fair comparison, all backbones in our study are pretrained on unlabeled ImageNet-1K \cite{Deng2009ImageNet}. The backbone is ResNet50. We evaluate the performance of CLLD approach on pretraining backbones for three lane detection algorithms: U-Net~\cite{Ronneberger2015UNet} ($\approx 28M parameters$), RESA~\cite{Zheng2021RESA}($\approx 25M parameters$), and CLRNet~\cite{Zheng_2022_CVPR} ($\approx 25M parameters$). 
All hyperparameters for the lane detection methods remain unchanged in our study. We employ the LARS optimizer , configured with a cosine learning rate schedule. The initial learning rate is set at 1.0. Additionally, we use a batch size of 1024 and a weight decay parameter set to 1e-5.
For the masking process, we set $\rho$ to 14.
In the momentum encoder, the momentum value starts at 0.99 and increases to 1.
We train ResNet50 on six Nvidia\textsuperscript{\textregistered} A100-40GB GPUs for 100 epochs. To study the effect of $\alpha$, we train ResNet50 with three different $\alpha$ values: 1, 2, and 3. 
We mask 30\% of the original image to generate the masked  version. We fine-tune all lane detection algorithms on two Nvidia\textsuperscript{\textregistered} RTX A6000 GPUs.}
\subsection{Benchmarks}
{
We evaluate CLLD on two lane detection benchmarks: CuLane~\cite{Pan2018SpatialCNN} and TuSimple~\cite{Shirke2019LaneDatasets}.
\subsubsection{CuLane} The CuLane includes 55 hours of video data, featuring both highways and urban scenarios.
CuLane includes nine validation subsets: \textit{Normal}, \textit{Crowd}, \textit{Night}, \textit{Noline}, \textit{Shadow}, \textit{Arrow}, \textit{Hlight}, and \textit{Curve}.
 Lane marking predictions are represented by 30-pixel wide lines. A prediction is considered a true positive if it has an Intersection over Union (IoU) greater than 0.5; values lower than 0.5 are classified as false positives. 
 In lane detection, a false positive refers to the incorrect identification of a lane where none exists, while a true negative indicates the correct recognition of the absence of a lane.
The absence of a prediction for a lane that is labeled in the ground truth is categorized as a False Negative (FP) while predicting lanes that do not exist is classified as a False Positive (FP).
Predictions categorized as FP and FN are considered unsuccessful.
The common metrics used for benchmarking~\cite{Pan2018SpatialCNN,Zheng_2022_CVPR,Zheng2021RESA} on CuLane are $Precision=\frac{TP}{TP+FP}$ and $Recall=\frac{TP}{TP+FN}$, and $F1-measure=\frac{2\times Precision\times Recall}{Precision+Recall}$.
\subsubsection{TuSimple}
The dataset comprises 3,626 videos, each with a resolution of 1280x720 pixels and an approximate length of 20 seconds.
\\
These videos are captured from a camera mounted on the windshield of a vehicle driving on highways under various weather and lighting conditions.
The accuracy metric used for benchmarking on TuSimple is defined as $accuracy=\frac{\Sigma_{clip} C_{clip}}{\Sigma_{clip} S_{clip}}$. Here, the correctly predicted points are denoted by $C_{clip}$, and the total number of lane points in ground truth is represented by $S_{clip}$.}
\section{Results}\label{section:results}
We specifically chose to evaluate CLLD with U-Net because it is a common encoder-decoder architecture used in various methods that treat lane detection as a semantic segmentation problem~\cite{Zou_2020_TVT}.
Additionally, we tested our method using RESA \cite{Zheng2021RESA}, which is currently the SOTA semantic segmentation lane detection and not based on the U-Net. Such independent validation is crucial to confirm the accuracy of our model.
Finally, we conduct an evaluation of CLLD using CLRNet~\cite{Zheng_2022_CVPR}, which is recognized as a leading anchor-based method for lane detection.
\subsection{Comparison with prior works}
\subsubsection{U-Net} The results of the lane detection using U-Net with CLLD pretraining, along with comparisons with other CL methods, are presented in \ref{tab:1}. More comprehensive results are available in Table 7 in the supplementary material.
The results indicate that CLLD with ($\alpha=3$) outperforms all other methods on the TuSimple benchmark.
Furthermore, according to most evaluation metrics, all CLLD versions outperform other methods on CuLane.
PixPro is the only method that offers better precision than CLLD; however, its recall is $\approx2 \%$ lower than the average recall achieved by CLLD.
Upon comparing CLLD with PixPro, it is observed that CLLD tends to generate a higher number of FP, whereas PixPro is more prone to producing FN.
This comparison indicates that CLLD excels in lane extrapolation compared to PixPro, whereas   PixPro demonstrates superior performance in lane interpolation.
It has been observed that all CLLD variants outperform VICRegL with a large margin despite being trained for 200 fewer epochs. This suggests that VICRegL, known for its trade-off between global and local features, might not be the most suitable choice for lane detection tasks.
We aim to accurately detect lanes not only under nominal scenarios but also under more challenging conditions with low visibility, such as scenarios where lane markings are obscured by shadows. 
As illustrated in \ref{tab:7}, CLLD suppresses all other CL methods, achieving an improvement of 7\% in such challenging scenarios.
{ CLLD markedly enhances lane detection performance in scenarios where the visibility of extensive lane segments is compromised by lighting conditions, such as shadows. This improvement is attributed to the maintenance of long-range dependencies between local features through the cross-similarity module. Consequently, lanes can be reconstructed by leveraging the similarity among illusion-invariant features, including edge and shape features. }
\begin{table}[ht]
  \caption{Performance of U-Net on CuLane and TuSimple with pretraining by different contrastive learning methods.}
  \label{tab:1}
  \centering
  \resizebox{\columnwidth}{!}{%
  \begin{tabular}{c|c | c c c|c}
    \hline
     \multirow{2}{*}{\textbf{Method}}&\multirow{2}{*}{\textbf{\#~Epoch}}&\multicolumn{3}{c|}{\textbf{CuLane}}&\textbf{TuSimple}\\
     \cline{3-6}
     &&Precision&Recall&F1-measure&Accuracy\\
      \hline
    PixPro \cite{Xie2021PropagateYourself}&100&\cellcolor{cyan!25}73.68&67.15&70.27&95.92\\
    VICRegL \cite{NEURIPS2022_39cee562}&300&67.75&63.43&65.54&93.58\\
    DenseCL \cite{Wang2021DenseCL}&200&63.8&58.4&60.98&96.13\\
    MoCo-V2 \cite{He2020CVPR}&200&63.08&57.74&60.29&96.04\\
      \hline
    CLLD ($\alpha=1$)&100&71.98&69.2&\cellcolor{cyan!25}70.56&95.9\\
    CLLD ($\alpha=2$)&100&70.69&69.36&70.02&95.98\\
    CLLD ($\alpha=3$)&100&71.31&\cellcolor{cyan!25}69.59&70.43&\cellcolor{cyan!25}96.17\\
   \hline
\end{tabular}}
\end{table}
\subsubsection{RESA} \ref{tab:2} (Appendix Table 8) illustrates the performance of RESA on CuLane and TuSimple with different contrastive learning methods for pretraining.
With the RESA architecture, all variations of CLLD surpass the performance of all other methods.
CLLD~($\alpha=1$) emerges as the best precision on the CuLane benchmark.
CLLD~($\alpha=3$) also outperforms other methods on the TuSimple benchmark and, for the most part, on CuLane, according to various evaluation metrics.
Similar to U-Net, the combination of RESA and CLLD shows a significant improvement ($\approx4\%$) on the shadow subset of CuLane.
This highlights the general enhancement in detecting lanes under low visibility conditions.
{ The behavior of CLLD on RESA can be understood through the same rationale applied to U-Net, suggesting a reinterpretation of its efficacy in enhancing lane detection under challenging lighting conditions.}
\begin{table}[ht]
  \caption{Performance of RESA \cite{Zheng2021RESA} on CuLane and TuSimple with different contrastive learnings.}
  \label{tab:2}
  \centering
  \resizebox{\columnwidth}{!}{%
  \begin{tabular}{c |c | c c c|c}
    \hline
     \multirow{2}{*}{\textbf{Method}}&\multirow{2}{*}{\textbf{\#~Epoch}}&\multicolumn{3}{c|}{\textbf{CuLane}}&\textbf{TuSimple}\\
     \cline{3-6}
     &&Precision&Recall&F1-measure&Accuracy\\
      \hline
    PixPro \cite{Xie2021PropagateYourself}&100&77.41&	73.69&	75.51&96.6\\
    VICRegL \cite{NEURIPS2022_39cee562}&300&76.27&	69.58&	72.77&96.18\\
    DenseCL \cite{Wang2021DenseCL}&200&77.67&	73.51&	75.53&96.28\\
    MoCo-V2 \cite{He2020CVPR}&200&78.12&	73.36&	75.66&96.56\\
      \hline
    CLLD ($\alpha=1$)&100&\cellcolor{cyan!25}79.01&	72.99&	75.88&96.74\\
    CLLD ($\alpha=2$)&100&78&	73.45&	75.66&96.78\\
    CLLD ($\alpha=3$)&100&78.34&	\cellcolor{cyan!25}74.29&	\cellcolor{cyan!25}76.26& \cellcolor{cyan!25}96.81\\
   \hline
\end{tabular}}
\end{table}
\subsubsection{CLRNet} Table~\ref{tab:3} (Appendix Table 9) presents the effectiveness of CLLD on CLRNet.
Compared to prior contrastive learning methods, CLLD achieves over $1\%$ improvement in recall for CuLane dataset.
It also achieves SOTA results on TuSimple accuracy and CuLane's F1-Measure.
Similar to previous studies, PixPro achieves better Precision on CuLane.
CLRNet is not a semantic segmentation approach. Instead, it detects lane anchors and connects them to achieve better lane extrapolation. 
{CLRNet exhibits marginal improvement with the integration of CLLD. This is attributed to CLRNet's strategy of refining lane detection at a higher layer; if it detects the majority of lane segments, it can extrapolate to fill in missing parts. However, if it fails to identify most of the lane segments, it may disregard the segments it has detected. While CLLD enhances the likelihood of detecting lane segments within CLRNet, the refinement of lanes at a higher level means that detecting discrete lane segments at lower levels may not significantly boost performance.}
\begin{table}[ht]
  \caption{Performance of CLRNet~\cite{Zheng_2022_CVPR} on CuLane and TuSimple with different pretraining strategies.}
  \label{tab:3}
  \centering
  \resizebox{\columnwidth}{!}{%
  \begin{tabular}{c|c | c c c|c}
    \hline
     \multirow{2}{*}{\textbf{Method}}&\multirow{2}{*}{\textbf{\#~Epoch}}&\multicolumn{3}{c|}{\textbf{CuLane}}&\textbf{TuSimple}\\
     \cline{3-6}
     &&Precision&Recall&F1-measure&Accuracy\\
      \hline
    PixPro\cite{Xie2021PropagateYourself}&100&\cellcolor{cyan!25}89.19&	70.39&	78.67&93.88\\
    VICRegL\cite{NEURIPS2022_39cee562}&300&87.72&	71.15&	78.72&89.01\\
    DenseCL\cite{Wang2021DenseCL}&200&88.07&	69.67&	77.8&85.15\\
    MoCo-V2\cite{He2020CVPR}&200&88.91&	71.02&	78.96&93.87\\
      \hline
    CLLD($\alpha=1$)&100&88.72&	71.33&	79.09&90.68\\
    CLLD($\alpha=2$)&100&87.95&	71.44&	78.84&93.48\\
    CLLD($\alpha=3$)&100&88.59&	\cellcolor{cyan!25}71.73&	\cellcolor{cyan!25}79.27&\cellcolor{cyan!25}94.25\\
   \hline
\end{tabular}}
\end{table}
\begin{table*}[t]
  \caption{Comparison of the performance of state-of-the-art lane detection methods on CuLane and TuSimple in two situations of pretraining with supervised learning and CLLD self-supervised learning. }
  \label{tab:4}
  \centering
  \resizebox{\textwidth}{!}{%
  \begin{tabular}{c| c|cc c| c c c c c c c c| c|c |cc c }
  \hline
  \multirow{3}{*}{\rotatebox{35}{\textbf{Method}}}&\multirow{3}{*}{\rotatebox{35}{\textbf{Pretrain}}}&\multicolumn{12}{c|}{\textbf{CuLane}}&\multicolumn{3}{c}{\textbf{TuSimple}}\\
    \cline {3-17}
    
    &&\multicolumn{3}{c|}{\textbf{Overall(\%)}}&\multicolumn{8}{c|}{\textbf{F1-Measure(\%)}}&\textbf{FP}&&&\\
    \cline {3-17}
    
    & &\textbf{Precision} & \textbf{Recall} & \textbf{F1} &\textbf{Normal}& \textbf{Crowd} & \textbf{Night} & \textbf{Noline}  & 
    \textbf{Shadow} & \textbf{Arrow} & \textbf{Hlight} & \textbf{Curve} &\textbf{Cross}&\textbf{Accuracy}&\textbf{FP}&\textbf{FN}\\
    \hline
    \multirow{2}{*}{U-Net~\cite{Ronneberger2015UNet} }&Supervised&70.93&69.65&70.28&89.82&67.72&64.95&40.49&68.13&84.48&59.83&67.02&2482&96.24&	0.0489&	0.0428\\
    &\cellcolor{cyan!25}CLLD&\cellcolor{cyan!25}71.31&\cellcolor{cyan!25}69.59&\cellcolor{cyan!25}70.43\cellcolor{cyan!25}&\cellcolor{cyan!25}89.8&\cellcolor{cyan!25}68.39&\cellcolor{cyan!25}64.65&\cellcolor{cyan!25}40.68&\cellcolor{cyan!25}68.86&\cellcolor{cyan!25}84.5&\cellcolor{cyan!25}58.93&\cellcolor{cyan!25}66.2&\cellcolor{cyan!25}2656&\cellcolor{cyan!25}96.17&	\cellcolor{cyan!25}0.055&	\cellcolor{cyan!25}0.045\\
    \hline
    \multirow{2}{*}{RESA\cite{Zheng2021RESA} }&Supervised&77.51&73.15&75.27&92.16&73.16&69.99&47.71&72.97&88.16&68.79&70.65&1503&96.67&0.031&0.0265\\
    &\cellcolor{cyan!25}CLLD&\cellcolor{cyan!25}78.34&	\cellcolor{cyan!25}74.29&	\cellcolor{cyan!25}76.26&	\cellcolor{cyan!25}92.57&	7\cellcolor{cyan!25}4.35&	\cellcolor{cyan!25}71.21&	\cellcolor{cyan!25}48.83&	\cellcolor{cyan!25}76.62&	\cellcolor{cyan!25}89.14&	\cellcolor{cyan!25}67.58&	\cellcolor{cyan!25}72.68&	\cellcolor{cyan!25}1454&\cellcolor{cyan!25}96.81&	\cellcolor{cyan!25}0.0343&	\cellcolor{cyan!25}0.0264\\
    \hline
    \multirow{2}{*}{CLRNet\cite{Zheng_2022_CVPR} }&Supervised&88.21&	71.88&	79.22&	93.1&	77.83&	74.3&	52.69&	76.92&	89.63&	73.16&	69.41&	1082&93.17&	0.0232&	0.0748\\
    &\cellcolor{cyan!25}CLLD&\cellcolor{cyan!25}88.59&	\cellcolor{cyan!25}71.73&	\cellcolor{cyan!25}79.27&	\cellcolor{cyan!25}92.94&	\cellcolor{cyan!25}77.44&	\cellcolor{cyan!25}74.43&	\cellcolor{cyan!25}53.3&	\cellcolor{cyan!25}81.2&	\cellcolor{cyan!25}89.31&	\cellcolor{cyan!25}72.46&	\cellcolor{cyan!25}68.4&	\cellcolor{cyan!25}1026&\cellcolor{cyan!25}94.25&	\cellcolor{cyan!25}0.214&	\cellcolor{cyan!25}0.069\\

  \hline
\end{tabular}}
\end{table*}

\subsection{Comparison with supervised learning}

Table~\ref{tab:4} presents the results of the CLLD pretraining strategy with supervised pretraining.
The best improvement ($\approx1\%$ in the average of all CuLane subsets), compared to supervised learning, is in the RESA with ResNet50 as the backbone.
CLLD also achieves a maximum $\approx4 \%$ increase in CuLane's low visible subsets, such as the shadow.
CLLD outperforms supervised learning on RESA for all metrics on both datasets, with the exception of the FP rate on TuSimple.
FP is the only metric for which supervised learning has provided better prediction outcomes than CLLD at a rate of $0.0343$ per prediction.

CLLD performance is equivalent to supervised learning based on most evaluation metrics on CLRNet {\scriptsize($\pm \le 1\%$)}.
For the shadow subset of CuLane, the accuracy of CLRNet was about $\approx 4\%$ better with CLLD pretraining than supervised learning.
CLLD performance in U-Net is comparable to supervised learning. 
It gains over $1\%$ better precision than supervised learning.
CLLD also produces over 300 more FPs than supervised learning in the cross subset of CuLane.
\subsection{Visual demonstration}
\ref{fig:fig5} illustrates a qualitative comparison of lane prediction preretained on CLLD compared with supervised learning and prior lane detection methods.
The results illustrate CLLD performance, especially for the most left lane with an occluder.
Most other training strategies detect lanes, but with many false positives, except DenseCL, which destroys the lane.
PixPro also has worse predictions than CLLD, with significant FP for the occluded part.
\begin{figure*}[ht]
    \centering
    \includegraphics[width=0.9\textwidth]{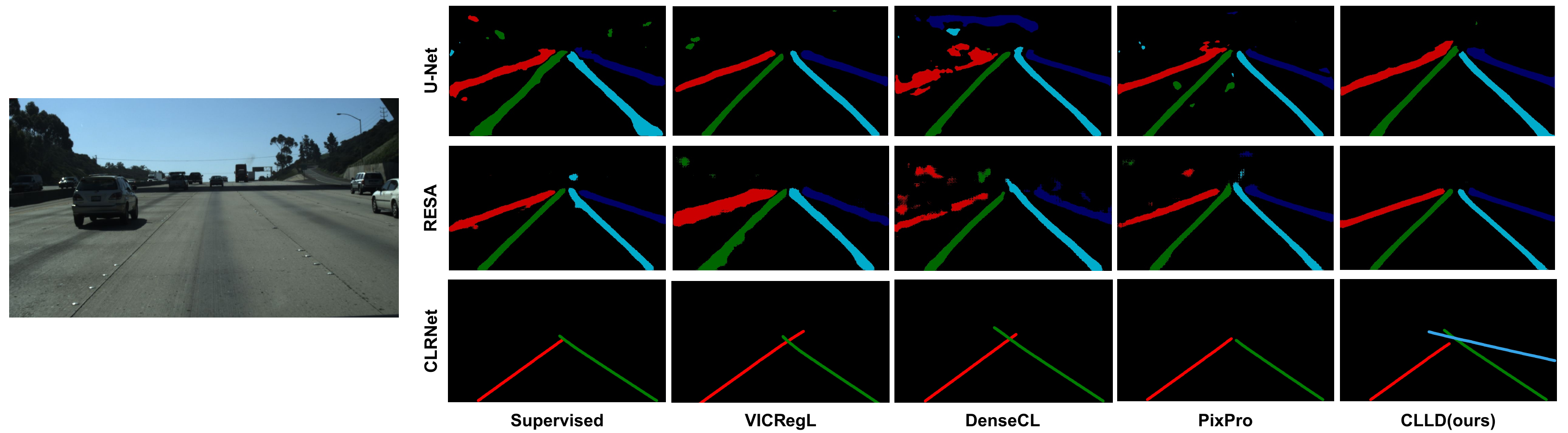}
    \caption{Qualitative comparison of the results of CLLD with prior SSL methods and supervised learning.}
    \label{fig:fig5}
\end{figure*}

\ref{fig:fig6} is the interpolated view of the latent layer on RESA for supervised and self-supervised learning (CLLD). 
The results show that supervised learning pays more attention to the texture of the road; however, CLLD focuses more on lanes' and objects' shapes, which is more important for lane detection.
Wu et al.~\cite{Wu2022VisionTransformers} study these differences in supervised and self-supervised learning behavior.
This may be a reason why self-supervised learning performs better on lane detection.
\begin{figure}[ht]

    \centering
    \includegraphics[width=0.8\columnwidth]{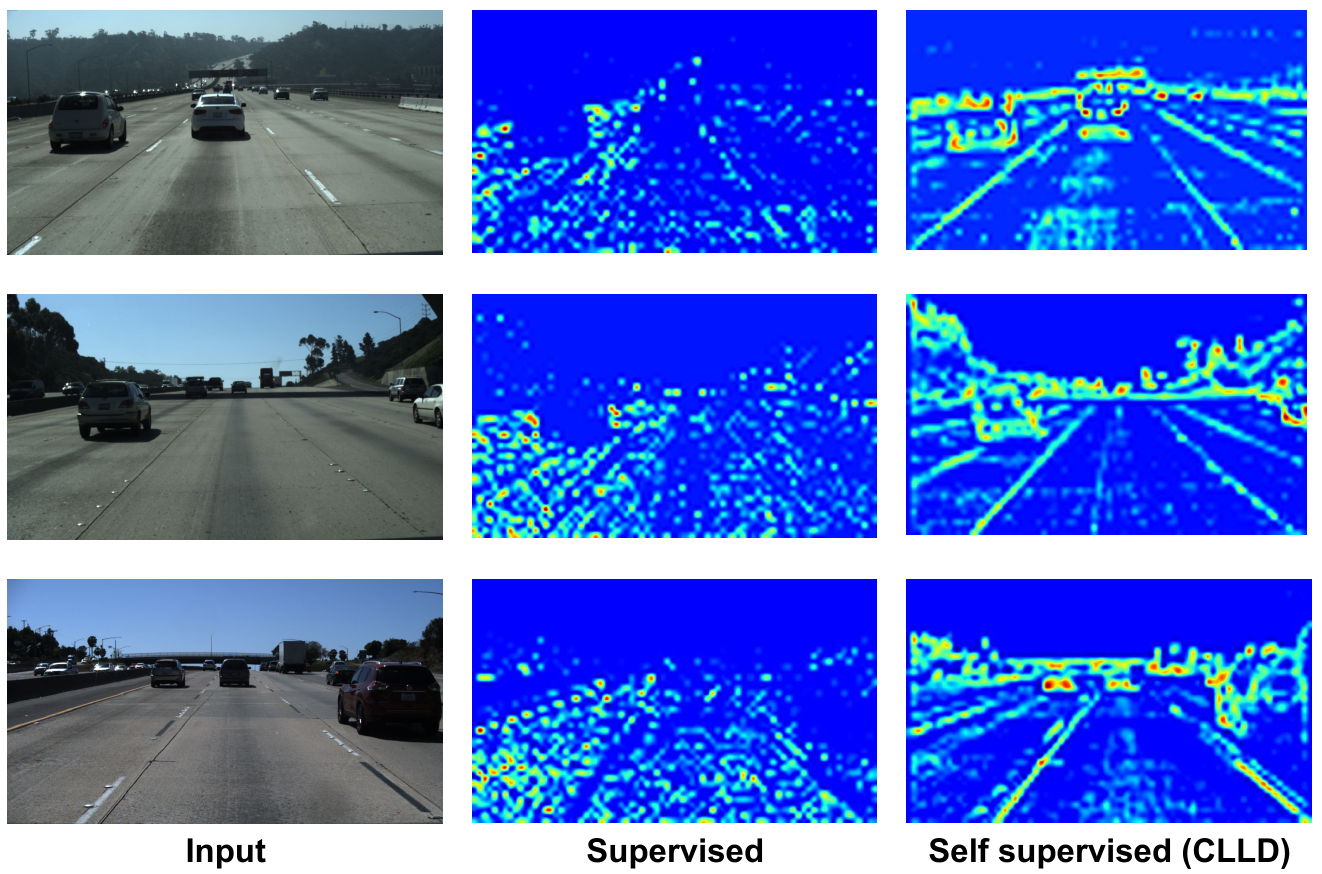}
    \caption{Low-level features in RESA; The left column is the input image, the middle column is low-level features in supervised learning, and the right column is low-level features in CLLD.}
    \label{fig:fig6}
\end{figure}
\section{Ablation study}\label{section:ablation}
\subsection{Similarity loss impact} \ref{tab:5} 
(Appendix Table 10) ablates CLLD performance with a single similarity loss, a single consistency loss, and the combination of them together. We did not study the absence of instance loss because it did not affect the segmentation results and used it for classification.
The accuracy of combining similarity and consistency is remarkably better than using only one.
$\mathcal{L}_{sim}+\mathcal{L}_{cons}$ performed significantly better results ($\approx5\%$) in challenging subsets of CULane dataset such as the shadow; however, the performance in the normal CuLane subset is not affected by the loss of similarity.
This observation illustrates the effect of similarity loss on the detection of low visible lanes.
\begin{table}[ht]
  \caption{\textbf{Ablation: Multi-task contrastive learning.} Comparison of the impact of similarity loss and consistency loss on the lane detection accuracy } 
  \label{tab:5}
  \centering
  \resizebox{0.8\columnwidth}{!}{%
  \begin{tabular}{c|c|c c c|c}
    \hline
    \multirow{2}{*}{$\mathcal{L}_{sim}$}&\multirow{2}{*}{$\mathcal{L}_{cons}$}&\multicolumn{3}{c|}{CuLane}&TuSimple\\
    \cline{3-6}
     &&Precision&Recall&F1&Accuracy\\
    \hline
    $*$&&76.91&70.82&73.74&95.94\\
    &$*$&77.41&	73.69&	75.51&95.92\\
    \hline
    $*$&	$*$&	78.34&	74.29&76.26	&96.17\\
  \hline
\end{tabular}}
\end{table}
\subsection{Impact of masking as the augmentation} 
\ref{tab:6} (Appendix Table 11) examines the effect of the masking strategy on overall accuracy. 
CLLD with masking yields a significantly better recall in CuLane ($\approx4\%$); however, it achieves a marginally lower precision ($\approx0.2\%$).
It increases the F-measure by an average of $\approx2\%$ over all CuLane subsets.
The combination with masking also improves the accuracy of TuSimple markedly ($\approx1\%$).
\begin{table}[ht]
  \caption{\textbf{Ablation: Impact of the masking as the augmentation.} Comparison of the accuracy of CLLD with and without using masking as the augmentation}
  \label{tab:6}
  \centering
  \resizebox{0.8\columnwidth}{!}{%
  \begin{tabular}{c|c c c|c}
    \hline
    \multirow{2}{*}{Masking}&\multicolumn{3}{c|}{CuLane}&TuSimple\\
    \cline{2-5}
     &Precision&Recall&F1&Accuracy\\
    \hline
    No&72.2&65.77&68.84&95.7\\
    Yes&71.98&69.2&70.56&96.17\\
  \hline
\end{tabular}}
\end{table}
\section{Discussion}

CLLD framework exhibits comparable performance in scenarios where lane delineations remain unaffected by occlusions or fading phenomena. However, it demonstrates a markedly superior performance in conditions where lane visibility is partially compromised, such as within the shadowed subsections of the CuLane dataset. This enhanced efficacy can be attributed to the incorporation of a cross-similarity module within the encoding process, which facilitates the capture of long-range dependencies across the visual field. Consequently, this mechanism affords the model the capability to infer the presence of lanes even in segments where they have become ostensibly invisible. It achieves this by leveraging the feature comparisons between visible lane segments and their occluded counterparts, thereby providing a probabilistic basis for the accurate prediction of the latter. Such an approach underscores the pivotal role of cross-similarity in enhancing the robustness of lane detection algorithms under varying visibility conditions.
Looking forward, we aim to explore the application of CLLD to Vision Transformers (ViTs) and develop a second version of CLLD compatible with both CNNs and ViTs. This expansion will address the evolving challenges in autonomous driving and lane detection technologies, potentially leading to even more robust lane detection systems.}
\section{Conclusion}
Our paper presents a novel self-supervised approach, Contrastive Learning for Lane Detection via cross-similarity (CLLD), designed to enhance the resilience of lane detection models in adverse conditions.
CLLD is a multi-task CL that addresses the challenge of detecting obscured lane markings caused by factors like poor lighting and weather by integrating our novel operation cross-similarity to local feature CLs.
{ CLLD captures long-range dependencies between different lane segments through the use of a cross-similarity operation. By computing the similarity between illusion-invariant features such as shape and edges, cross-similarity can reconstruct lane patterns even in segments with low visibility. This approach utilizes similarity to enhance lane detection precision for lanes that are partially occluded or have invisible due to natural variations.}

Our method (CLLD) demonstrates remarkable improvements over existing contrastive learning techniques, particularly excelling in scenarios with low visibility, such as shadows. 
 In the future, our focus will be on identifying challenges associated with applying CLLD to Vision Transformers (ViTs) and developing a second version of CLLD that is compatible with both CNNs and ViTs.
\subsubsection{Acknowledgement}
This work was supported in part by This study was co-funded by the European Union and Estonian Research Council via project TEM-TA138, the Swedish Innovation Agency VINNOVA project AutoDeep. The computations were enabled by resources provided by the supercomputing resource Berzelius provided by National Supercomputer Centre at Link\"oping
University and the Knut and Alice Wallenberg foundation.
\bibliography{aaai24}

\begin{table*}[ht]
  \caption{U-Net complete table }
  \label{tab:7}
  \centering
  \resizebox{\textwidth}{!}{%
  \begin{tabular}{c| c| c |c c c| c c c c c c c c|c |c|c c }
  \hline
  \multirow{3}{*}{\rotatebox{35}{\textbf{Method}}}&\multirow{4}{*}{\rotatebox{35}{\textbf{Epochs}}}&\multirow{3}{*}{\rotatebox{35}{\textbf{Backbone}}}&\multicolumn{12}{c|}{\textbf{CuLane}}&\multicolumn{3}{c}{\textbf{TuSimple}}\\
    \cline {4-6}
    \cline{7-14}
    \cline{15-18}
    &&&\multicolumn{3}{c|}{\textbf{Overall(\%)}}&\multicolumn{8}{c|}{\textbf{F1-Measure(\%)}}&\textbf{FP}&&&\\
    \cline {4-6}
    \cline{7-14}
    \cline{15-18}
    
    &&& \textbf{Precision} & \textbf{Recall} & \textbf{F1} &\textbf{Normal}& \textbf{Crowd} & \textbf{Night} & \textbf{Noline}  & 
    \textbf{Shadow} & \textbf{Arrow} & \textbf{Hlight} & \textbf{Curve} &\textbf{Cross}&\textbf{Accuracy}&\textbf{FP}&\textbf{FN}\\
    \hline
    
    PixPro&100&ResNet50&\cellcolor{cyan!25}73.68&67.15&70.27&88.27&\cellcolor{cyan!25}68.85&\cellcolor{cyan!25}65.54&41.12&57.25&83.17&58.2&66.07&\cellcolor{cyan!25}1820&95.92&	0.0643&	0.052\\
    VICRegL&300&ResNet50&67.75&63.43&65.54&85.56&63.63&60.15&35.52&55.54&78.17&50.67&60.47&2673&93.58&	0.1427&	0.1077\\
    MoCo-V2&200&ResNet50&63.08&57.74&60.29&80.02&61.91&56.51&33.54&53.3&74.42&49.13&61.07&7561&96.13&	\cellcolor{cyan!25}0.0524&	0.047\\
    DenseCL&200&ResNet50&63.8&58.4&60.98&80.65&63.01&56.84&35.17&53.98&74.73&48.28&61.5&7784&96.04&	0.04742&	0.0505\\
    \hline

    CLLD(ours)($\alpha=1$)&100&ResNet50&71.98&69.2&\cellcolor{cyan!25}70.56&89.74&68.23&65.14&40.82&64.72&83.5&59.46&65.62&2112&\cellcolor{cyan!25}96.17&	0.055&	\cellcolor{cyan!25}0.045\\
    CLLD(ours)($\alpha=2$)&100&ResNet50&70.69&69.36&70.02&89.29&67.68&64.36&40.04&\cellcolor{cyan!25}70.96&83.33&\cellcolor{cyan!25}59.93&64.97&2117&95.98&	0.0564&	0.0488\\

    CLLD(ours)($\alpha=3$)&100&ResNet50&71.31&\cellcolor{cyan!25}69.59&70.43&\cellcolor{cyan!25}89.8&68.39&64.65&\cellcolor{cyan!25}40.68&68.86&\cellcolor{cyan!25}84.5&58.93&\cellcolor{cyan!25}66.2&2656&95.9&	0.0636&	0.0515\\

  \hline
\end{tabular}}
\end{table*}

\begin{table*}[ht]
\vspace{0.7cm}
  \caption{RESA complete table }
  \label{tab:8}
  \centering
  \resizebox{\textwidth}{!}{%
  \begin{tabular}{c| c| c |c c c| c c c c c c c c|c |c|c c }
  \hline
  \multirow{3}{*}{\rotatebox{35}{\textbf{Pretrain}}}&\multirow{4}{*}{\rotatebox{35}{\textbf{Epochs}}}&\multirow{3}{*}{\rotatebox{35}{\textbf{Backbone}}}&\multicolumn{12}{c|}{\textbf{CuLane}}&\multicolumn{3}{c}{\textbf{TuSimple}}\\
    \cline {4-6}
    \cline{7-14}
    \cline{15-18}
    &&&\multicolumn{3}{c|}{\textbf{Overall(\%)}}&\multicolumn{8}{c|}{\textbf{F1-Measure(\%)}}&\textbf{FP}&&&\\
    \cline {4-6}
    \cline{7-14}
    \cline{15-18}
    
    &&& \textbf{Precision} & \textbf{Recall} & \textbf{F1} &\textbf{Normal}& \textbf{Crowd} & \textbf{Night} & \textbf{Noline}  & 
    \textbf{Shadow} & \textbf{Arrow} & \textbf{Hlight} & \textbf{Curve} &\textbf{Cross}&\textbf{Accuracy}&\textbf{FP}&\textbf{FN}\\
    \hline

    PixPro&100&ResNet50&77.41&	73.69&	75.51&	92.78&	73.12&	70.17&	48.72&	71.89&	\cellcolor{cyan!25}89.55&	66.68&	72.02&	1699
&96.6&	0.0362&	0.0298\\
    VICRegL&300&ResNet50&76.27&	69.58&	72.77&	90.27&	71.56&	68.39&	44.27&	59.88&	85.75&	63.05&	67.75&	1933
&96.18&	0.044&	0.041\\
    MoCo-V2&200&ResNet50&78.12&	73.36&	75.66&	92.45&	74.19&	70.35&	47.47&	71.27&	88.31&	\cellcolor{cyan!25}69.23&	69.12&	1471
&96.56&\cellcolor{cyan!25}	0.0341&	0.0307\\
    DenseCL&200&ResNet50&77.67&	73.51&	75.53&	92.69&	73.33&	70.51&	48.14&	70.36&	89.03&	65.13&	70.45&	1456
&96.28&	0.04117&	0.04112\\
    \hline

    CLLD(ours)($\alpha=1$)&100&ResNet50&\cellcolor{cyan!25}79.01&	72.99&	75.88&	92.49&	74.19&	70.8&	46.42&	75.17&	89.15&	67.07&	70.4&	\cellcolor{cyan!25}1276
&96.74&	\cellcolor{cyan!25}0.0341&	0.0277\\
    CLLD(ours)($\alpha=2$)&100&ResNet50&78	&73.45&	75.66&	92.47&	73.85&	70.45&	47.41&	75.88&	88.67&	65.93&	70.28&	1584
&96.78&	0.037&	\cellcolor{cyan!25}0.0262\\

    CLLD(ours)($\alpha=3$)&100&ResNet50&78.34&	\cellcolor{cyan!25}74.29&	\cellcolor{cyan!25}76.26&	\cellcolor{cyan!25}92.57&	\cellcolor{cyan!25}74.35&	\cellcolor{cyan!25}71.21&	\cellcolor{cyan!25}48.83&	\cellcolor{cyan!25}76.62&	89.14&	67.58&	\cellcolor{cyan!25}72.68&	1454
&\cellcolor{cyan!25}96.81&	0.0343&	0.0264\\

  \hline
\end{tabular}}
\vspace{0.9cm}
\end{table*}

\begin{table*}[ht]
\vspace{0.7cm}
  \caption{CLRNet complete table }
  \label{tab:8}
  \centering
  \resizebox{\textwidth}{!}{%
  \begin{tabular}{c| c| c |c c c| c c c c c c c c|c |c|c c }
  \hline
  \multirow{3}{*}{\rotatebox{35}{\textbf{Pretrain}}}&\multirow{4}{*}{\rotatebox{35}{\textbf{Epochs}}}&\multirow{3}{*}{\rotatebox{35}{\textbf{Backbone}}}&\multicolumn{12}{c|}{\textbf{CuLane}}&\multicolumn{3}{c}{\textbf{TuSimple}}\\
    \cline {4-6}
    \cline{7-14}
    \cline{15-18}
    &&&\multicolumn{3}{c|}{\textbf{Overall(\%)}}&\multicolumn{8}{c|}{\textbf{F1-Measure(\%)}}&\textbf{FP}&&&\\
    \cline {4-6}
    \cline{7-14}
    \cline{15-18}
    
    &&& \textbf{Precision} & \textbf{Recall} & \textbf{F1} &\textbf{Normal}& \textbf{Crowd} & \textbf{Night} & \textbf{Noline}  & 
    \textbf{Shadow} & \textbf{Arrow} & \textbf{Hlight} & \textbf{Curve} &\textbf{Cross}&\textbf{Accuracy}&\textbf{FP}&\textbf{FN}\\
    \hline

    PixPro&100&ResNet50&\cellcolor{cyan!25}89.19&	70.39&	78.67&	91.69&	77.34&	74.45&	51.8&	80.52&	88.26&	72.99&	68.23&	\cellcolor{cyan!25}898&93.88&	0.0355&	0.0667\\
    VICRegL&300&ResNet50&87.72&	71.15&	78.72&	92.57&	75.94&	\cellcolor{cyan!25}74.53&	52.52&	79.45&	88.85&	\cellcolor{cyan!25}73.88&	67.57&	1070&89.01&	0.3115&	0.2316\\
    MoCo-V2&200&ResNet50&88.91&	71.02&	78.96&	93&	77.35&	73.6&	51.56&	78.91&	\cellcolor{cyan!25}89.77&	71.35&	68.61&	1011&93.87&	0.11276&	0.10391\\
    DenseCL&200&ResNet50&88.07&	69.67&	77.8&	91.83&	75.93&	72.97&	50.28&	77.64&	88.39&	71.67&	67.41&	983&85.15&	0.1497&	0.2714\\
    \hline

    CLLD(ours)($\alpha=1$)&100&ResNet50&88.72&	71.33&	79.09&	92.85&	77.6&	74.07&	52.39&	80.42&	89.15&	72.29&	67.1&	989&90.68&	0.0217&	0.1179\\
    CLLD(ours)($\alpha=2$)&100&ResNet50&87.95&	71.44&	78.84&	92.85&	76.93&	74.17&	51.96&	81.02&	89.07&	72.15&	\cellcolor{cyan!25}69.5&	1186&93.48&	0.0352&	0.0749\\

    CLLD(ours)($\alpha=3$)&100&ResNet50&88.59&	\cellcolor{cyan!25}71.73&	\cellcolor{cyan!25}79.27&	\cellcolor{cyan!25}92.94&	\cellcolor{cyan!25}77.44&	74.43&	\cellcolor{cyan!25}53.3&	\cellcolor{cyan!25}81.2&	89.31&	72.46&	68.4&	1026&\cellcolor{cyan!25}94.25&	\cellcolor{cyan!25}0.214&	\cellcolor{cyan!25}0.069\\

  \hline
\end{tabular}}
\vspace{0.9cm}
\end{table*}
\begin{table*}[ht]
  \caption{\textbf{Ablation: Impact of the loss function.} Comparison of CLLD accuracy with or without using Similarity loss (Complete table)}
  \label{tab:9}
  \centering
  \resizebox{\textwidth}{!}{%
  \begin{tabular}{c|c|c c c|cccccccc|c|c|ccc}
    \hline
    \multirow{3}{*}{$\mathcal{L}_{sim}$}&\multirow{3}{*}{$\mathcal{L}_{cons}$}&\multicolumn{12}{c|}{CuLane}&\multicolumn{3}{c}{TuSimple}\\
    \cline{3-17}
    &&\multicolumn{3}{c|}{Overall(\%)}&\multicolumn{8}{c|}{F-Measures(\%)}&FP\\
    \cline{3-17}
     &&Precision&Recall&F1&Normal&	Crowd&	Night&	Noline&	Shadow&	Arrow&	Hlight&	Curve&	Cross&Accuracy&FP&FN\\
    \hline
    $*$&&76.91&70.82&73.74&91.67&	71.21&	66.84&	46.55&	70.56&	87.1&	66.43&	66.75&	1571&95.94&	0.04521&	0.0449\\
    &$*$&77.41&	73.69&	75.51&92.78&	73.12&	70.17&	48.72&	71.89&	89.55&	66.68&	72.02&	1699&95.92&	0.0643&	0.052\\
    \hline
    $*$&	$*$&	78.34&	74.29&	76.26&92.57&	74.35&	71.21&	48.83&	76.62&	89.14&	67.58&	72.68&	1454&96.17&	0.055&	0.045\\
  \hline
\end{tabular}}
\end{table*}

\begin{table*}[ht]
  \caption{\textbf{Ablation: Impact of RandomDrop.} Comparison of CLLD accuracy with or without using RandomDrop as the augmentation (Complete table)}
  \label{tab:10}
  \centering
  \resizebox{\textwidth}{!}{%
  \begin{tabular}{c|c c c|cccccccc|c|c|ccc}
    \hline
    \multirow{3}{*}{RandomDrop}&\multicolumn{12}{c|}{CuLane}&\multicolumn{3}{c}{TuSimple}\\
    \cline{2-16}
    &\multicolumn{3}{c|}{Overall(\%)}&\multicolumn{8}{c|}{F-Measures(\%)}&FP\\
    \cline{2-16}
     &Precision&Recall&F1&Normal&	Crowd&	Night&	Noline&	Shadow&	Arrow&	Hlight&	Curve&	Cross&Accuracy&FP&FN\\
    \hline
    No&	72.2&	65.77&	68.84&	87.34&	67.95&	63.61&	38.37&	61.25&	81.17&	56.41&	62.95&	2324&95.7&	0.0652&	0.0554\\
    Yes&	71.98&	69.2&	70.56&	89.74&	68.23&	65.14&	40.82&	64.72&	83.5&	59.46&	65.62&	2112&		96.17&	0.055&	0.045\\
  \hline
\end{tabular}}
\end{table*}

\end{document}